%% file: main.tex
\documentclass[11pt,a4paper]{article}
\usepackage[]{cogsci}

\cogscifinalcopy 

\usepackage[hidelinks]{hyperref}

\usepackage{pslatex}
\usepackage{apacite}
\usepackage{float} 

\usepackage{times}
\usepackage{soul}
\usepackage{url}
\usepackage{clrscode}
\usepackage{multirow}
\usepackage[utf8]{inputenc}
\usepackage[small]{caption}
\usepackage{graphicx}
\usepackage{amsmath}
\usepackage{amsthm}
\usepackage{booktabs}
\usepackage{algorithm}
\usepackage{algorithmic}
\usepackage{comment}
\usepackage{array}
\newcolumntype{x}[1]{>{\centering\arraybackslash\hspace{0pt}}p{#1}}

\newcommand{\blank}{$\rule{.5cm}{0.15mm}$ \ }
\usepackage{array}
\newcolumntype{P}[1]{>{\centering\arraybackslash}p{#1}}
\newcolumntype{M}[1]{>{\centering\arraybackslash}m{#1}}

\renewcommand{\paragraph}[1]{\vspace{2mm}\noindent\textbf{#1}\hspace{2mm}}

\title{Probing Neural Language Models for Human Tacit Assumptions}
  
\author{{\large \bf Nathaniel Weir, Adam Poliak, and Benjamin Van Durme }\\
  Department of Computer Science, Johns Hopkins University \\
  \texttt{\{nweir,azpoliak,vandurme\}@jhu.edu} }

\begin{document}

\maketitle
\begin{abstract}
Humans 
carry 
\emph{stereotypic tacit assumptions} (STAs)~\cite{prince_function_1978},
or propositional beliefs
about generic concepts. Such 
associations are crucial for 
understanding
natural
language.
We construct a diagnostic set of word prediction prompts  
to evaluate 
whether recent neural contextualized language models trained on large text corpora
capture STAs. Our prompts are based on human responses in
a psychological study of conceptual associations.
We find 
models
to be profoundly effective at retrieving concepts given associated properties.
Our results demonstrate empirical evidence that stereotypic conceptual representations are captured in neural models 
derived from semi-supervised linguistic exposure.

\end{abstract}

\input{sections/1_introduction.tex}
\input{sections/2_background.tex}
\input{sections/3_features2concept.tex}
\input{sections/4_concept2feature.tex}

\input{sections/5_capturing_prince.tex}
\input{sections/6_discussion.tex}
\bibliographystyle{apacite}

\setlength{\bibleftmargin}{.125in}
\setlength\bibitemsep{0pt}
\setlength{\bibindent}{-\bibleftmargin}

\bibliography{zotero}
\newpage
\input{sections/7_appendix}
\end{document}

%% file: sections/1_introduction.tex
\section{Introduction}
Recognizing generally accepted properties about concepts is key to understanding natural language~\cite{prince_function_1978}.
For example, if one mentions a bear, one does not have to explicitly 
describe the animal as having teeth or claws, or as being a predator or a threat. 
This phenomenon 
 reflects one's held stereotypic tacit assumptions (STAs), 
i.e. propositions commonly attributed to
``classes of entities''~\cite{prince_function_1978}.
STAs, a form of common knowledge~\cite{walker_common_1991}, 
are salient to 
 cognitive scientists
 concerned with how human representations of knowledge and meaning manifest. 

As ``studies in norming responses are prone to repeated responses across subjects''~\shortcite{poliak_hypothesis_2018}, 
cognitive scientists demonstrate empirically that humans share assumptions about properties associated with concepts~\shortcite{mcrae_semantic_2005}.
We take these conceptual assumptions as one instance of STAs and ask whether recent contextualized language models trained on large text corpora capture them. 
In other words, do models correctly distinguish concepts associated with a given set of properties?  
To answer this question, we 
design 
fill-in-the-blank diagnostic tests (\autoref{fig:iter-prompting}) based on existing data of 
concepts with corresponding sets of human-elicited properties. 

 \begin{figure}[t!]
    \centering
    \small
    \begin{tabular}{
    p{.55\columnwidth}p{.3\columnwidth}}
\toprule
\textbf{Prompt} & \textbf{Model Predictions} \\
\midrule
\emph{A \blank has fur.} & dog, cat, fox, ...  \\[1mm]
\emph{A \blank has fur, is big, and has claws.} & cat, \textbf{bear}, lion, ... \\[1mm]
\emph{A \blank has fur, is big, has claws, has teeth, is an animal, eats, is brown, and lives in woods.} & \textbf{bear}, wolf, cat, ... \\
\bottomrule
    \end{tabular}
    \caption{
    The concept \textbf{bear} as a target emerging as the highest ranked predictions
    of the neural LM \proc{RoBERTa-L} (Liu et al., 2019)  
    when prompted with conjunctions of the concept's
    human-produced properties. 
    }
    \label{fig:iter-prompting}
\end{figure}
By tracking conceptual recall from prompts of iteratively concatenated conceptual properties,
we find that  
the popular neural language models,
\proc{BERT}~\shortcite{devlin_bert_2019} and \proc{RoBERTa}~\shortcite{liu_roberta_2019},
capture STAs. 
We observe that
\proc{RoBERTa} consistently outperforms \proc{BERT} in correctly associating concepts with their defining properties
across multiple metrics; this performance discrepancy is consistent with many other language understanding tasks~\shortcite{wang_glue_2019}. 
We also find that models associate concepts with perceptual categories of properties (e.g. visual) worse than with non-perceptual ones  (e.g. encyclopaedic or functional). 

We further examine whether STAs can be extracted \textit{from} the models by designing prompts akin to those shown to humans in psychological studies~\shortcite{mcrae_semantic_2005,devereux_centre_2013}. We find significant overlap between model and human responses, but with notable differences.
We provide qualitative examples in which the models' predictive associations differ from humans', yet are still sensible given the prompt. Such results 
highlight the difficulty of constructing word prediction prompts that
elicit particular forms of reasoning from models optimized purely to predict co-occurrence.

Unlike other work analyzing linguistic meaning captured in sentence representations derived from language models~\shortcite{conneau_what_2018,tenney_what_2019}, 
we do not fine-tune the models to perform
any task; we instead find that the targeted tacit assumptions ``fall out'' purely from semi-supervised masked language modeling.
Our results demonstrate that exposure to large corpora alone, without multi-modal perceptual signals or task-specific training cues, may enable a model to sufficiently capture STAs. 

%% file: sections/2_background.tex
\section{Background}

\paragraph{Contextualized Language Models}
Language models (LMs) assign probabilities to sequences of text. They are trained on large text corpora to predict the probability of a new word based on its 
surrounding context.
Unidirectional models approximate for any 
text sequence  $\boldsymbol{w} = [w_1, w_2, \dots w_N]$ the factorized left-context probability  ${p(w) = \prod^N_{i=1} p(w_i \mid w_{1}\dots w_{i-1})}$. 
Recent neural \textit{bi-directional} language models 
estimate the probability of an intermediate `masked out' token given both left and right context; this task is colloquially ``masked language modelling'' (MLM).
Training in this way produces a probability model that, given input sequence $\boldsymbol{w}$ and an arbitrary vocabulary word 
predicts the distribution ${p(w_i = v \mid w_1, \dots w_{i-1},  w_{i+1},\dots w_{n})}$.
When neural bi-directional LMs trained for MLM 
are subsequently used as contextual encoders,\footnote{That is, when used to obtain contextualized representations of words and sequences.} performance across a wide range of language understanding tasks greatly improves.

We investigate two recent neural LMs:
\textbf{\underline{B}}i-directional \textbf{\underline{E}}ncoder \textbf{\underline{R}}epresentations from \textbf{\underline{T}}ransformers (\proc{BERT})~\shortcite{devlin_bert_2019}
 and 
\underline{\textbf{R}}obustly \underline{\textbf{o}}ptimized \underline{\textbf{BERT}} \underline{\textbf{a}}pproach (\proc{RoBERTa})~\shortcite{liu_roberta_2019}.
In addition to the MLM objective,
\proc{BERT} is trained with an auxiliary objective of next-sentence prediction. 
\proc{BERT} is trained on 
a book corpus
and English Wikipedia.
Using an identical neural architecture, \proc{RoBERTa} is trained for purely MLM (no next-sentence prediction) on a much larger dataset with words masked out of larger input sequences. Performance increases ubiquitously on standard NLU tasks when \proc{BERT} is replaced with \proc{RoBERTa} as an off-the-shelf contexual encoder.

\paragraph{Probing 
Language Models via Word Prediction}
Recent research employs word prediction tests to explore whether contextualized language models 
capture a range of 
linguistic phenomena, e.g. 
syntax~\cite{goldberg_assessing_2019}, 
pragmatics, semantic roles, and negation~\cite{ettinger_what_2019}.
These diagnostics have psycholinguistic origins; they draw an analogy between
the ``fill-in-the-blank'' word predictions of a pre-trained language model and
distribution of aggregated human responses in cloze tests designed to target specific sentence processing phenomena.
Similar tests have been used to evaluate how well these models capture 
symbolic reasoning~\shortcite{talmor_olmpics_2019}
and relational facts~\cite{petroni_language_2019}.

\paragraph{Stereotypic Tacit Assumptions}
Recognizing associations between concepts and their defining properties is key to natural language understanding and plays  
``a critical role in language both for the conventional meaning of utterances, and in conversational inference''~\cite{walker_common_1991}.
\textit{Tacit assumption} (TAs) are commonly accepted beliefs about specific entities (\textit{Alice has a dog}) 
and \textit{stereotypic} TAs (STAs) pertain to a generic concept, or a class of entity (\textit{people have dogs})~\cite{prince_function_1978}. 
While held by individuals, STAs are generally agreed upon and are vital for reflexive reasoning and pragmatics; Alice might tell Bob `I have to walk my dog!,' but she 
does not need to say
``I am a person, and people have dogs, and dogs need to be walked, so I have to walk my dog!''
Comprehending STAs allows for generalized recognition of new categorical instances, and facilitates learning \textit{new} categories~\shortcite{lupyan_language_2007}, 
as shown in early word learning by children~\shortcite{hills_categorical_2009}. 
STAs are not explicitly facts.\footnote{E.g., ``countries have presidents'' does not apply to \textit{all} countries.} Rather, they are sufficiently 
probable assumptions to be associated with concepts by a majority of people. A partial inspiration for this work was the observation by \citeA{extracting-implicit-knowledge-from-text} that the concept attributes most supported by peoples' \textit{search engine query logs}~\cite{pasca_you_2007} were strikingly similar to examples of STAs listed by Prince. That is, there is strong evidence that the beliefs people hold about particular conceptual attributes (e.g. ``countries have kings''), are reflected in the aggregation of their most frequent search terms (``what is the name of the king of France?'').

Our goal is to determine whether contextualized language models exposed to large corpora encode associations between concepts and their tacitly assumed properties. 
We develop probes that specifically test a model's ability to recognize STAs. Previous works~\shortcite{rubinstein_how_2015, sommerauer_firearms_2018,da_cracking_2019} have tested for similar types of stereotypic beliefs; they use supervised training of probing classifiers~\shortcite{conneau_what_2018} to identify concept/attribute pairs. In contrast, our word prediction diagnostics find that these associations \textit{fall out} of semi-supervised LM pretraining. In other words, the neural LM inducts STAs as a byproduct of learning co-occurrence without receiving explicit cues to do so.

%% file: sections/3_features2concept.tex
\section{Probing for Stereotypic Tacit Assumptions} 

\begin{figure*}[t!]
    \centering
\includegraphics[width=.48\textwidth]{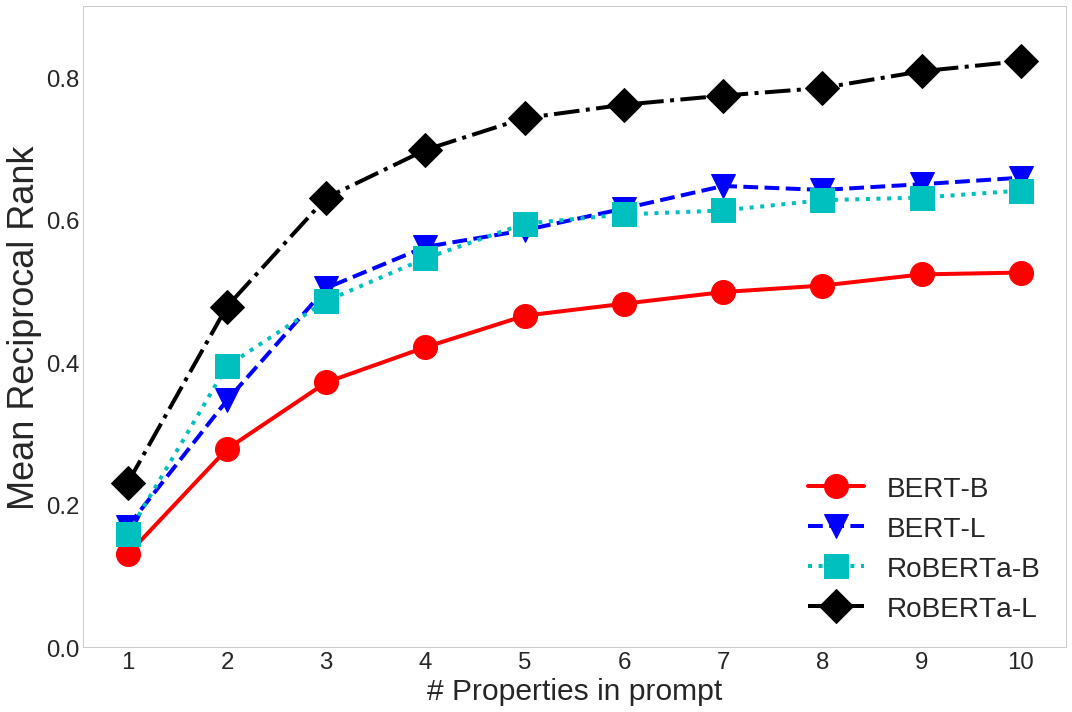}
\includegraphics[width=.48\textwidth]{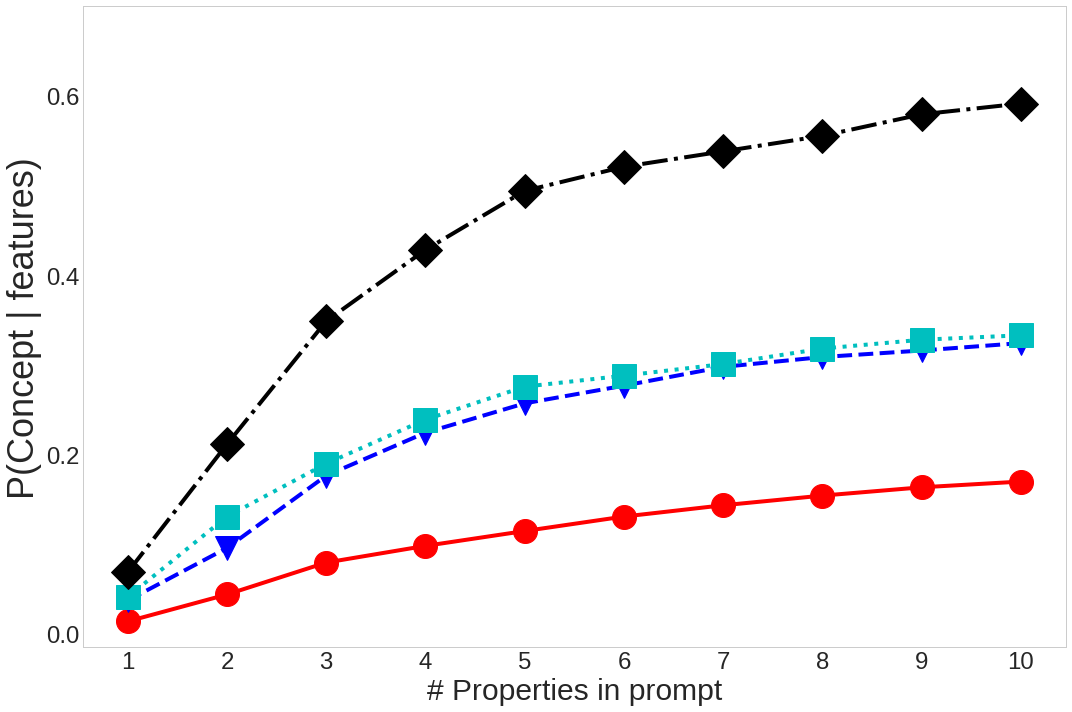}\\
    
    \caption{Results from neural LM concept retrieval diagnostic. Mean reciprocal rank and assigned probability of correct concept word sharply increase with the number of conjunctive properties in the prompt.}
    \label{fig:features2concept}
\end{figure*}
Despite introducing the notion of STAs,  
\citeA{prince_function_1978} provides only a few examples. 
We therefore draw from other literature to create diagnostics that evaluate how well a contexualized language model captures the phenomenon.
Semantic feature production norms,  i.e. properties elicited from human subjects regarding generic concepts,
fall under the category of STAs.
Interested in determining ``what people know about different things in the world,''\footnote{
Wording taken from instruction shown to participants---as shown in Appendix B of \citeA{mcrae_semantic_2005}}
 \citeA{mcrae_semantic_2005} had human subjects 
 list properties that they associated with individual concepts.
When many people individually attribute the same properties to a specific concept, collectively they provide STAs. We target the
 elicited properties that were most often repeated across the subjects. 
 
\paragraph{Prompt Design }
We
construct 
prompts 
for evaluating 
STAs in LMs
by leveraging
the CSLB Concept Property Norms~\cite{devereux_centre_2013}, a large extension
of the McRae 
study that contains $638$ concepts each linked with roughly $34$ associated properties.
The fill-in-the-blank prompts are natural language statements 
in which the 
target concept associated with a set of human-provided properties is the
missing word
. 
If LMs accurately predict the missing concept, we posit that they 
encode the given STA set.
We iteratively grow prompts by appending conceptual properties into a single compound verb phrase (\autoref{fig:iter-prompting}) until the verb phrase contains $10$ properties.
Since we test for $266$ concepts, this process creates a total of $2,660$ prompts.\footnote{Because LMs are highly sensitive to the `a/an' determiner preceding a masked word e.g. 
LMs far prefer to complete \textit{``\textbf{A} \blank buzzes,''} with ``bee,'' but prefer e.g. ``insect'' to complete \textit{``\textbf{An} \blank buzzes.''}, a task issue noted by \citeA{ettinger_what_2019}.
We remove examples containing concepts that begin with vowel sounds. A prompt construction that simultaneously accepts words that start with both vowels and consonants is left for future work.}
\citeA{devereux_centre_2013} record production frequencies (PF) enumerating how many people produced
each property for a given concept.
For each concept, we select and append the properties with the highest PF in decreasing order. 
Iteratively growing prompts
enables a \textit{gradient of performance} - we observe concept retrieval given few ``clue'' properties and track improvements as more are given. 

\vspace{2mm}
\noindent\textbf{Probing Method} \hspace{2mm}
Prompts are fed as toknized sequences to the neural LM encoder with the concept token replaced with a \texttt{[MASK]}. 
A softmax is taken over the 
final hidden vector
extracted from the model at the index of the masked token to obtain a probability distribution over the vocabulary of possible words.
 Following \citeA{petroni_language_2019}, we use a pre-defined, case-sensitive vocabulary
 of roughly $21$K 
 tokens to control for the possibility that a model's vocabulary size influences its rank-based performance.\footnote{The vocabulary is constructed from the unified intersection of those used to train \proc{BERT} and \proc{RoBERTa}. We omit concepts that are not contained within this intersection.} 
We use this probability distribution to obtain a ranked list of words that the model believes should be the missing $t$ token.
We evaluate the \proc{base} (\proc{-B}) and \proc{large} (\proc{-L}) cased models of  \proc{BERT} and \proc{RoBERTa}.

\begin{figure*}[t!]
    \centering
(a)\includegraphics[width=.48\textwidth]{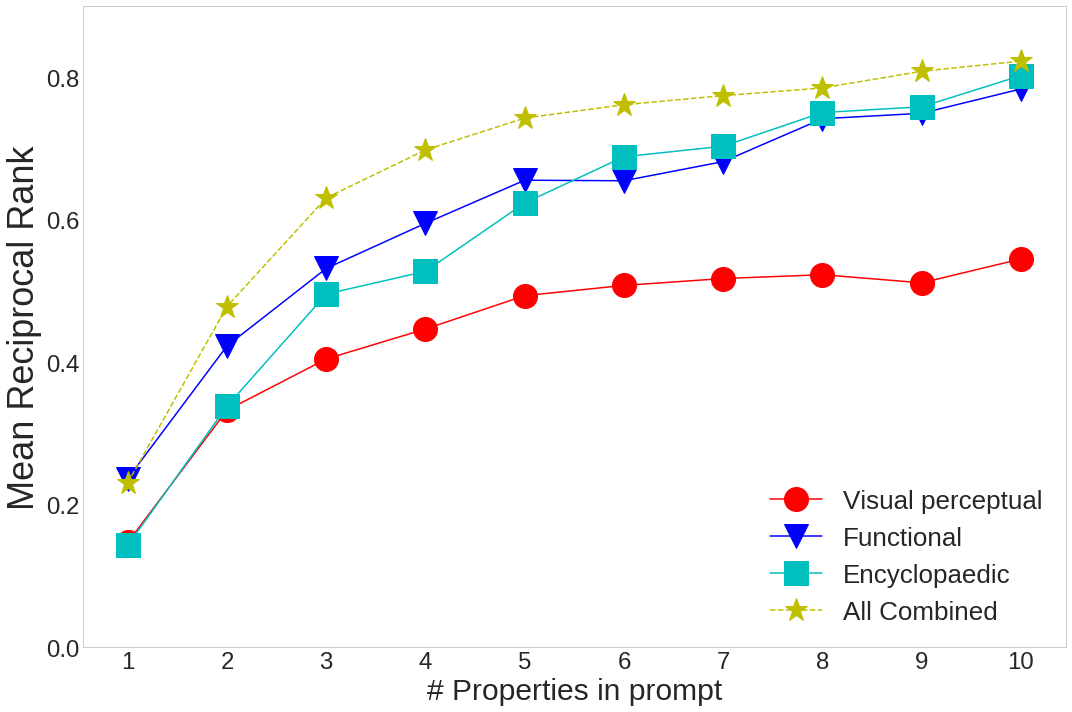}(b)\includegraphics[width=.48\textwidth]{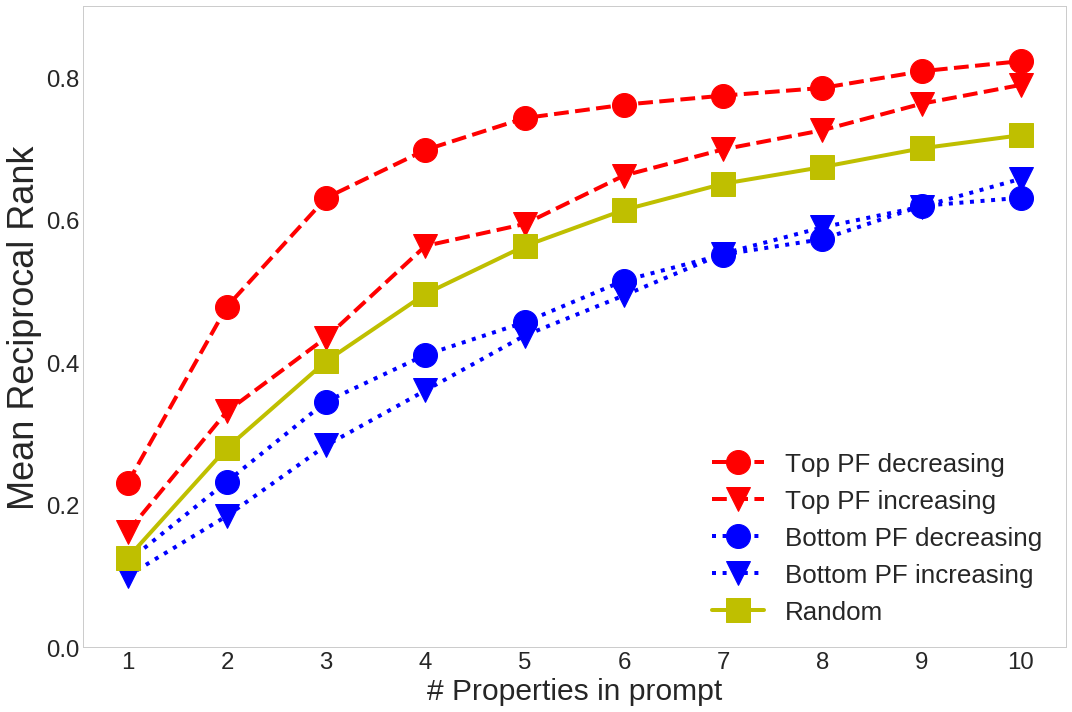}

    \caption{(a) Comparison of \proc{RoBERTa-L}'s  performance
 given only properties from each category versus all combined.
 (b) \proc{RoBERTa-L} performance given the property sets with the top vs bottom production frequencies (PF) ordered in increasing vs decreasing PF. Plotted against a randomly sampled and reordered baseline.}
    \label{fig:features2concept-followup}
\end{figure*}
\paragraph{Evaluation Metrics}
We use mean reciprocal rank (MRR), or $1 / \text{rank}_{\text{LM}}(\text{target concept})$, 
a metric more sensitive to fine-grained differences in rank than other common retrieval metrics such as recall.
We 
track the predicted rank of a target concept from relatively low ranks given few `clue' properties to much higher ranks as more properties are appended.
MRR above $0.5$ for a test set indicates that a model's top 1 prediction is correct
in a majority of examples.
We also report the overall probability the LM assigns to the target concept regardless of rank. This allows us to measure model \textit{confidence} beyond empirical task performance.

\section{Results} 

 \autoref{fig:features2concept} displays the results.
  When given just one property, \proc{RoBERTa-L} achieves a MRR of $0.23$, indicating that the target concept appears on average in the model's top-5 fill-in predictions (over the whole vocabulary). 
  The increase in MRR and model confidence (y-axis) as properties
 are iteratively appended to prompts (increasing x-axis) demonstrates that the LMs more accurately retrieve the missing concept when given more associated properties.
  MRR steeply increases for all models as properties are added to a prompt, 
  but we find less stark improvements after the first four or five.
   The \proc{large} models consistently outperform their \proc{base} variants under both metrics, as do 
 \proc{RoBERTa}s over the \proc{BERT}s of the same size. \proc{RoBERTa-B} and \proc{BERT-L} perform interchangeably.
Notably, \proc{RoBERTa-L} achieves a higher performance on both metrics when given just $4$ `clue' properties than any other model when provided with all $10$.
\proc{RoBERTa-L} assigns \textit{double} the target probability at $10$ properties than that of the next best model (\proc{RoBERTa-B}). 
Thus, \proc{RoBERTa-L} is profoundly more confident in its \textit{correct} answers than any other model.
However, that all models achieve at least between .5 and .85 MRR conditioned on $10$ properties illustrates the effectiveness of all considered models in identifying concepts given STA sets.

\paragraph{Qualitative Analysis}
Examples of prompts and corresponding model predictions are shown in Appendix \autoref{tab:feature2concept-qualitative}.
We find that model predictions are nearly always
grammatical and semantically 
sensible. 
Highly-ranked incorrect answers generally apply to a subset of the conjunction of properties, or are correct at an intermediate iteration but become precluded by subsequently appended properties.~\footnote{E.g. \textit{tiger} and \textit{lion} are correct for \textit{`A \blank has fur, is big, and has claws'} but reveal to be incorrect with the appended \textit{`lives in woods'}}
We note that an optimal performance may not be perfect; not all prompts uniquely identify the target concept, even at $10$ properties.~\footnote{E.g. the properties of \textit{buffalo} do not distinguish it from \textit{cow}.
}
However, 
models still perform nearly as well as could be expected given the ambiguity.

\paragraph{Properties Grouped by Category} 
To measure whether the the \textit{type} of property affects the ability of LMs to retrieve a concept,
we create additional prompts that only contain properties of specific categories
as grouped by \citeA{devereux_centre_2013}: visual perceptual (bears have fur), 
functional (eat fish), 
and encyclopaedic (are found in forests).\footnote{We omit the categories ``other perceptual'' (bears growl) and  ``taxonomic'' (bears are animals), as few concepts have more than 2-3 such properties.}

\autoref{fig:features2concept-followup}a shows that \proc{RoBERTa-L} performs interchangeably well given just encyclopedic or functional type properties.
\proc{BERT} (not shown) shows a similar overall pattern, but it performs slightly better given encyclopedic properties than functional. 
Perceptual properties 
are overall less helpful for models to distinguish concepts than non-perceptual.
This may be the product of category specificity; while perceptual properties are produced by humans nearly as frequently as non-perceptual, the average perceptual property is assigned to nearly twice as many CSLB concepts as the average non-perceptual (6 to 3). 
However, the empirical finding 
coheres with previous conclusions that models that learn from language alone lack knowledge of perceptual features~\shortcite{collell_is_2016,lucy_are_2017}.

\paragraph{Selecting and Ordering Prompts}
When designing the probes, we selected and appended the $10$ properties with the highest production frequencies (PF) in decreasing order. 
To investigate whether these selection and ordering choices affect a 
model's performance in the retrieval task, 
we compare the top-PF property selection method with an alternative selection criterion using 
the \textit{bottom}-PF properties. For both selection methods, we compare the decreasing-PF ordering with a reversed, increasing-PF order.  
We compare the resulting 4 evaluations against a random baseline that measures performance using a random permutation of a randomly-selected set of properties.\footnote{
The random baseline's performance is averaged over 5 random permutations of 5 random sets for each concept.} 

\autoref{fig:features2concept-followup}b compare the differences in performance. Regardless of ordering, the selection of the top (bottom)-PF features improves (reduces) model performance relative to the random baseline. Ordering by decreasing PF improves performance over the opposite direction by up to $0.2$ for earlier sizes of property conjunction, but the two strategies converge in performance for larger sizes.
This indicates that the selection and ordering criteria of the properties may matter
when adding them to prompts.
The properties with lower PF are correspondingly less beneficial for model performance. This suggests that assumptions that are less stereotypic---that is, highly salient to fewer humans---are less well captured by the LMs.

%% file: sections/4_concept2feature.tex
\section{Eliciting Properties from Language Models}
We have found that neural language models capture to a high degree the relationship between human-produced sets of stereotypic tacit assumptions and their associated concepts. 
Can we use the LMs 
to \textit{retrieve} the conceptual properties 
under the same type of setup used for human elicitation?
We design prompts to replicate the ``linguistic filter''~\cite{mcrae_semantic_2005}  through which the human subjects conveyed conceptual assumptions.

In the human elicited studies, subjects were asked to list properties that would complete ``\{concept\} \{relation\}...'' prompts in which the relation could 
take on\footnote{Selected at the discretion of the subject via a drop-down menu.}
one of four fixed phrases: \textit{is}, \textit{has}, \textit{made of}, and \textit{does}.
We mimic this protocol using the first three relations\footnote{We do not investigate the \textit{does} relation or the open-ended \textit{``...''} relation, because the resulting human responses are not easily comparable with LM predictions using template-based prompts. We construct prompts using \textit{is a} and \textit{has a} for broader dataset coverage.}
and compare the properties predicted by the LMs to 
the corresponding human response sets. Examples of this protocol are shown in \autoref{tab:concept2feature_ex}.

\paragraph{Comparing LM Probabilities with Humans
}
We can consider the listed properties as samples from a 
fuzzy notion of 
a human STA \textit{distribution} 
conditioned on the concept and relation.
These STAs reflect how humans codify their probabilistic beliefs about the world. What a subject writes down about the `dog' concept reflects what that subject believes from their experience to be sufficiently ubiquitous, i.e. extremely probable, for all `dog' instances.
The dataset also portrays a distribution \textit{over} listed STAs. Not all norms are produced by all participants given the same concepts and relation prompts; this reflects how individuals hold different sets of STAs about the same concept.
Through either of these lenses, 
we can speculate that the human subject produces the sample e.g. `fur' from some $p(\text{STA} \mid \text{concept } = \text{\textit{bear}},
\ \text{relation} = \textit{has})$.~\footnote{This formulation should be taken with a grain of salt; the subject is given all relation phrases at once and has the opportunity to fill out as many (or few) completions as she deems salient, provided that in combination there are at least 5 total properties listed.}
We can consider our protocol to be sampling from a LM approximation of such a conditional distribution.
\begin{table}[!t]
    \centering
    \small
    \begin{tabular}{p{.25\columnwidth}c m{.5cm}cm{.5cm}}
    \toprule
         \multirow{2}{*}{\textbf{Context}} & \multicolumn{2}{c}{ \textbf{Human}} & 
    \multicolumn{2}{c}{\textbf{\proc{RoBERTa-L}}}  \\ \cmidrule(lr){2-3} \cmidrule(lr){4-5}
                                      & \textbf{Response} & \textbf{PF} & \textbf{Response} & $p_{\text{LM}}$  \\ \midrule
    \multirow[t]{6}{.24\columnwidth}{\textit{(Everyone knows that) a bear has \blank.}} & fur & 27 & teeth & .36\\
                                              & claws & 15 & claws & .18 \\
                                              & teeth & 11 & eyes & .05\\
                                              & cubs & 7 & ears & .03 \\
                                              & paws & 7 & horns & .02 \\ \midrule
    \multirow[t]{5}{.24\columnwidth}{\textit{(Everyone knows that) a ladder is made of \blank.}}& metal & 25 & wood & .33\\
                                              & wood & 20 & steel & .08\\
                                              & plastic & 4 & metal & .07\\
                                              & aluminum & 2 & aluminum & .03\\
                                              & rope & 2 & concrete & .03\\
    \bottomrule
    \end{tabular}
\caption{Example concept/relation prompts with resulting 
human
and \proc{RoBERTa-L} responses (and corresponding production frequencies and LM probabilities, resp.). Portions of context prompts encased in () were only shown to the model, not human.}
    \label{tab:concept2feature_ex}
\end{table}

\paragraph{Limits to Elicitation}
Asking language models to list properties via word prediction is inherently limiting,
as the models are not primed to specifically produce 
\textit{properties} beyond whatever cues we can embed in the context of a sentence.  In contrast, human subjects were asked directly ``What are the properties of X?''~\cite{devereux_centre_2013}. This is a highly semantically constraining question
that cannot be directly asked of an off-the-shelf language model.

The phrasing of the question to humans also has 
implications regarding salience:
when describing a dog, humans would rarely, if never, describe a dog as being \textit{``larger than a pencil''},
 even though humans are ``capable of verifying'' this property~\cite{mcrae_semantic_2005}.
 Even if they do produce a property as opposed to an alternative lexical completion, it may be unfair to expect language models to replicate how human subjects prefer to list properties that distinguish and are salient to a concept (e.g. \textit{`goes moo'}) as opposed to listing properties that apply to many concepts (e.g. \textit{`has a heart'}). 
Thus, comparing properties elicited by language models to those elicited by humans is a challenging endeavour.  
Anticipating this issue, we prepend the phrase `Everyone knows that' to our  prompts.
They 
therefore take the form 
shown in the left column of \autoref{tab:concept2feature_ex}.
For the sake of comparability, we evaluate the models' responses against only the human responses that fit the same syntax. We also remove human-produced properties with multiple words following the relation (e.g. `\textit{is} \underline{found in forests}') since the contextualized LMs under consideration can only predict a single missing word. 
Our method produces a set of between 495 and 583 prompts for each of the relations considered.

\begin{table}[!t]
\centering
\small
\begin{tabular}{lclrrrr}
\toprule
   \textbf{Relation}     & \textbf{$|\text{Data}|$} &     \textbf{Metric}        &     \textbf{Bb} &     \textbf{Bl} &     \textbf{Rb} &     \textbf{Rl} \\
\midrule 
\textbf{is} & 583 & \textbf{$\text{mAP}_{\proc{vocab}}$} &  .081 &  .080 &  .078 &  \textbf{.190 }\\
        && \textbf{$\text{mAP}_{\proc{sens}}$} &  .131 &  .132 &  .105 &  \textbf{.212} \\
        && \textbf{$\rho_{\text{Human PF}}$} &  .062 &  .100 &  .062 &  \textbf{.113} \\\midrule
\textbf{is a} & 506 &  \textbf{$\text{mAP}_{\proc{vocab}}$} &  .253 &  .318 &  .266 &  \textbf{.462} \\
        && \textbf{$\text{mAP}_{\proc{sens}}$} &  .393 &  .423 &  .387 &  \textbf{.559} \\
        && \textbf{$\rho_{\text{Human PF}}$} &  .226 &  \textbf{.389} &  .385 &  .386 \\\midrule
\textbf{has} & 564 & \textbf{$\text{mAP}_{\proc{vocab}}$} &  .098 &  .043 &  .151 &  \textbf{.317} \\
        && \textbf{$\text{mAP}_{\proc{sens}}$} &  .171 &  .138 &  .195 &  \textbf{.367} \\
        && \textbf{$\rho_{\text{Human PF}}$} &  .217 &  .234 &  .190 &  \textbf{.316} \\\midrule
\textbf{has a} & 537 &  \textbf{$\text{mAP}_{\proc{vocab}}$} &  .202 &  .260 &  .136 &  \textbf{.263} \\
        && \textbf{$\text{mAP}_{\proc{sens}}$} &  .272 &  .307 &  .208 &  \textbf{.329} \\
        && \textbf{$\rho_{\text{Human PF}}$} &  .129 &  .153 &  .174 &  \textbf{.209} \\\midrule
\textbf{made of} & 495 &  \textbf{$\text{mAP}_{\proc{vocab}}$} &  .307 &  .328 &  .335 & \textbf{ .503 }\\
        && \textbf{$\text{mAP}_{\proc{sens}}$} &  .324 &  .339 &  .347 &  \textbf{.533} \\
        && \textbf{$\rho_{\text{Human PF}}$} &  .193 &  .182 &  .075 &  \textbf{.339} \\
\bottomrule
\end{tabular}

\caption{Mean average precision and Spearman $\rho$ with human PF for LM prediction of properties given concept/relation pairs. \textbf{B} and \textbf{R} indicate \proc{BERT} and \proc{RoBERTa}, \textbf{b} and \textbf{l} indicate \proc{-base} and \proc{-large}.}
\label{tab:concept2feature-results}
\end{table}

\paragraph{Results} 
We use the 
information retrieval metric mean average precision (mAP)
for ranked sequences of predictions in which there are multiple correct answers.
We define mAP here given $n$ test examples:
\[
    \text{mAP} = \frac{1}{n} \sum^n_{i=1}
 \sum^{|\text{vocab}|}_{j=1}P_i(j)\Delta r_i(j)
    \]
where $P_i(j)$ = precision@$j$  and $\Delta r_i(j)$ is the change in recall from item $j-1$ to $j$ for example $i$.
We report mAP on prediction ranks over a LM's entire vocabulary ($\text{mAP}_{\proc{vocab}}$), but also over a much smaller vocabulary ($\text{mAP}_{\proc{sens}}$) comprising the set of human completions
that fit the given prompt syntax \textit{for all concepts in the study}.  
This follows the intuition that responses given for a set of concepts 
are likely \textit{not} attributes of 
the other concepts, and models should be sensitive to this discrepancy.
While mAP measures the ability to distinguish the \textit{set}\footnote{Invariant to order of correct answers.} of correct responses from incorrect responses, 
we also evaluate probability assigned \textit{among} the correct answers by computing average Spearman's $\rho$ between human production frequency and LM probability.

Results using these metrics are displayed in \autoref{tab:concept2feature-results}.
We find that \proc{RoBERTa-L} outperforms the other models by up to double mAP. 
No model's rank ordering of correct answers correlates particularly strongly with human production frequencies.
When we narrow the models' vocabulary to include only the property words produced by humans for a given syntax, we find that performance ($\text{mAP}_{\proc{sens}}$) increases ubiquitously.

\paragraph{Qualitative Analysis} 
Models generally provide coherent and grammatically acceptable completions. Most outputs fall under the category of `verifiable by humans,' which as noted by McRae et al. could be listed by humans given sufficient instruction. We observe properties that apply to the concept but are not in the dataset~\footnote{E.g. `\textit{hamsters are real}' and `\textit{motorcycles have horsepower}'.} and properties that apply to senses of a concept that were not considered in the human responses.~\footnote{While human subjects list only properties of the object \textit{anchor}, LMs also provide properties of a television anchor.}
We find that some prompts are not sufficently syntactically constraining, and license non-nominative completions. The relation \textit{has} permits past participle completions (e.g. `has arrived') along with the targeted nominative attributes (`has wheels').  
We also find that models idiosyncratically
favor specific words regardless of the concept, which can lead to unacceptable completions.\footnote{\proc{RoBERTa-B} often blindly produces `\textit{has legs}', the two \proc{BERT} models predict that nearly all concepts are `\textit{made of wood},' and all models except \proc{RoBERTa-L} often produce `\textit{is dangerous}.'} 
We provide examples predictions produced by models in Appendix \autoref{tab:concept2feature-failures}.

\paragraph{Effect of Prompt Construction} We investigate the extent to which our choice of lexical framing impacts model performance by ablating the step in which ``everyone knows that'' is prepended to the prompt.
We find a relatively wide discrepancy in effects; with the lessened left context, models perform on average $.05$ and $.1$ mAP worse on the \textit{is} and \textit{has} relations respectively, but perform  $.06$ and $.01$ mAP \textul{better} on \textit{is a} and \textit{has a}. Notably, \proc{RoBERTa-L} sees a steep drop in performance on the \textit{has} relation, losing nearly .3 mAP. 
We observe that models exhibit highly varying levels of instability given the choice of context. This
highlights the difficulty in constructing prompts that effectively target the same type of lexical response from any arbitrary bi-directional LM.

%% file: sections/5_capturing_prince.tex
\begin{table}[!t]
    \centering
    \small
    \begin{tabular}{p{.43\columnwidth}p{.43\columnwidth}} 
    \toprule
    \textbf{Prince Example}  &  \proc{Roberta-L} \\ \midrule 
    A \textbf{\textul{person}} has parents, siblings,
relatives, a home,
a pet, a car,
a spouse, a job. ,
&  person [.73], child [.1], human [.04], family [.03], kid [.02]  \\ \midrule
A country has
a \textbf{\textul{leader}}, a \textbf{\textul{duke}},
\textbf{\textul{borders}}, a \textbf{\textul{president}},
a \textbf{\textul{queen}}, \textbf{\textul{citizens}},
\textbf{\textul{land}}, a \textbf{\textul{language}},
and a \textbf{\textul{history}}. &  constitution [.23], history [.07], culture [.07], soul [.04], budget [.03], border [.03], leader [.03], currency [.02], population [.02]\\
    \bottomrule
    \end{tabular}
    \caption{\proc{RoBERTa-L} captures Prince's own exemplary STAs (target completions bolded), as shown by predictions of both concept and properties (associated probability in brackets).}
    \label{fig:prince}
\end{table}
\section{Capturing Prince's STAs}

We return to \citeA{prince_function_1978} to investigate whether neural language models,
which we have found to capture STAs elicited from humans by McRae, 
do so as well for what she envisioned.
Prince lists some of her \textit{own} STAs about the concepts \textit{country} and \textit{person}. 
We apply the methodologies of the previous experiments and show the resulting conceptual recall and feature productions in \autoref{fig:prince}. We find significant overlap in both directions of prediction. Thus, the exact examples of  
basic information about the world that Prince considers core to  
discourse and language processing are clearly captured by the neural LMs under investigation.

%% file: sections/6_discussion.tex
\section{Conclusion}
We have explored whether the notion owing to \citeA{prince_function_1978} of the stereotypic tacit assumption (STA), a type of background knowledge core to natural language understanding, is captured by contexualized language modeling.
We developed diagnostic experiments derived from human subject responses to a psychological study of conceptual representations
and observed that recent contextualized LMs trained on large corpora may indeed capture such important information. 
Through word prediction tasks akin to human cloze tests, our results provide a lens of quantitative and qualitative exploration of whether \proc{BERT} and \proc{RoBERTa} capture concepts and associated properties. 
We illustrate that the conceptual knowledge elicited from humans by \citeA{devereux_centre_2013} is indeed contained within an encoder; when a speaker may mention something that \textit{`flies'} and \textit{`has rotating blades,'} the LM can infer the description is of a \textit{helicopter}.
We hope that our work serves to further research in exploring 
the extent of semantic and linguistic knowledge captured by contextualized language models.

\section{Acknowledgements}
This work was supported in part by DARPA KAIROS (FA8750-19-2-0034). The views and conclusions contained in this work are those of the authors and should not be interpreted as representing official policies or endorsements of DARPA or the U.S. Government.

%% file: sections/7_appendix.tex
\input{tables/feature2concept_examples}

\input{tables/concept2feature_failures}
\input{tables/pasca_attributes}
\section{Appendix}
The following tables show qualitative results of our experiments. \autoref{tab:feature2concept-qualitative} shows \proc{BERT-L} and \proc{RoBERTa-L}'s predicted concepts with associated log probabilities given iteratively longer conjunctions of human-elicited properties. \autoref{tab:concept2feature-failures} shows examples of property production given concept/relation prompts; they are chosen as notable failure cases that exhibit shortcomings of the elicitation and evaluation protocol.

\subsection{Connection to Web-Extracted Class Attributes}
This work shows that neural contextualized LMs encode concept/property pairs commonly held among people as hypothesized by \citeA{prince_function_1978}. Their ubiquity is reflected in the frequency by which they were produced by subjects of the CSLB property norms study.
\citeA{pasca_you_2007}, also concerned with concepts and their attributes, show that these pairs are reflected in the logs of people's web searches. Their work, which proposes automatic concept/attribute extraction based on frequency of occurrence in web logs, can be viewed as additional support for Prince's STAs; people hold the beliefs that concepts have particular attributes (e.g. ``countries have kings''), and then reflect such beliefs in their queries (``who is the king of France?'').

As such, we examine whether the neural LMs under investigation capture a sample of the concept/attribute sets documented in \citeA{pasca_you_2007}. Results shown in \autoref{fig:pasca} show the significant degree to which these sets are captured by \proc{Roberta-L}.

%% file: tables/feature2concept_examples.tex
\begin{figure*}
    \centering
    \scriptsize
    \begin{tabular}{p{.24\textwidth}p{.34\textwidth}p{.34\textwidth}}
    \toprule
    \textbf{Context} & \proc{BERT-L} & \proc{RoBERTa-L}  \\ \midrule
    \textit{A \underline{\textbf{bus}} has wheels.}& 
car [-2.4], wheel [-2.9], wagon [-3.2], horse [-3.3]
, vehicle [-3.9]
& 
car [-1.8], \textbf{bus [-1.9]}, train [-2.4], bicycle [-2.6]
, horse [-3.4]   
\\ 
\textit{A \underline{\textbf{bus}} has wheels, is made of metal, carries, has a driver, is red, and transports people.} & 
car [-1.6], cart,[-2.1], \textbf{bus [-2.1]}, truck,[-2.7]
, wagon[-2.9]
& 
\textbf{bus [-0.6]},
car [-1.7],
train [-2.7],
cab [-3.6]
, taxi [-3.7]
\\ 
\textit{A \underline{\textbf{bus}} has wheels, is made of metal, carries, has a driver, is red, transports people, has seats, is transport, is big, and has windows.}     & car [-1.1], \textbf{bus [-1.5]}, truck [-2.6], vehicle [-3.0], tram [-3.2] &       \textbf{bus [-0.8]}, car [-0.9], train [-3.2], truck [-3.6], vehicle [-3.9] \\
\midrule
\textit{A \underline{\textbf{cake}} is tasty.} &    bite [-3.1], meal [-3.3], duck [-3.7], little [-3.9], steak [-4.0] &             lot [-3.8], steak [-4.1], meal [-4.6], pizza [-4.6], duck [-4.8] \\ 
\textit{A \underline{\textbf{cake}} is tasty, is eaten, is made of sugar, is made of flour, and is made of eggs.}   &    \textbf{cake [-2.4]}, dish [-3.2], sweet [-3.6], pie [-3.8], dessert [-3.9] &       cookie [-1.2], \textbf{cake [-1.2]}, pie [-2.8], meal [-2.9], banana [-3.4] \\
\textit{A \underline{\textbf{cake}} is tasty, is eaten, is made of sugar, is made of flour, is made of eggs, has icing, is baked, is sweet, is a kind of pudding, and is for special occasions.} &         \textbf{cake [-.7]}, pie [-3.0], dessert [-3.1], jam [-4.0], dish [-4.1] &     \textbf{cake [-.1]}, pie [-2.6], cookie [-3.9], dessert [-4.3], cream [-6.5] \\
\midrule
\textit{A \underline{\textbf{buffalo}} has horns.} &         lion [-2.9], horse [-3.3], goat [-3.6], man [-3.6], bull [-3.9] &        bull [-2.8], wolf [-2.9], horse [-3.0], goat [-3.1], cow [-3.3] \\
\textit{A \underline{\textbf{buffalo}} has horns, is hairy, is an animal, is big, and eats grass.}    &        goat [-2.6], man [-2.7], horse [-3.1], bear [-3.3], lion [-3.5] &        bull [-1.6], cow [-1.8], lion [-2.4], goat [-2.4], horse [-3.1] \\
\textit{A \underline{\textbf{buffalo}} has horns, is hairy, is an animal, is big, eats grass, lives in herds, is a mammal, is brown, eats, and has four legs.}    &     man [-1.8], person [-2.0], goat [-2.9], human [-3.3], horse [-3.3] &         cow [-1.1], lion [-2.3], bear [-2.5], deer [-2.6], bull [-2.6] \\
\midrule
\textit{A \underline{\textbf{tiger}} has stripes.} &        number [-4.2], line [-4.2], stripe [-4.3], lot [-4.7], color [-4.8] &        \textbf{tiger [-2.4]}, dog [-3.4], cat [-3.6], lion [-3.7], bear [-3.7] \\
\textit{A \underline{\textbf{tiger}} has stripes, is a cat, is orange, is big, and has teeth.} &   cat [-1.1], \textbf{tiger [-2.5]}, dog [-2.6], person [-3.1], man [-3.6] &        \textbf{tiger [-.5]}, cat [-1.9], lion [-2.8], dog [-3.7], bear [-3.7] \\
\textit{A \underline{\textbf{tiger}} has stripes, is a cat, is orange, is big, has teeth, is black, is endangered, is a big cat, is an animal, and is a predator.} &  cat [-.4], \textbf{tiger [-2.7]}, person [-3.5], lion [-4.2], dog [-4.3] &        cat [-.3], \textbf{tiger [-1.6]}, lion [-3.5], fox [-4.4], bear [-4.5] \\
\midrule
\textit{A \underline{\textbf{book}} has pages.} &  
page [-0.9],
\textbf{book [-1.2]},
file [-3.8],
chapter [-4.1],
word [-4.5] &
\textbf{book [-0.3]},
diary [-2.2],
novel [-2.8],
journal [-3.8],
notebook [-3.8] \\
\textit{A \underline{\textbf{book}} has pages, is made of paper, has a cover, is read, and has words.} &   \textbf{book  [-0.06]},
novel  [-4.7],
manuscript  [-4.7],
Bible  [-5.4],
dictionary  [-5.5] &
\textbf{book  [-0.01]},
novel  [-4.8],
newspaper  [-6.0],
dictionary  [-6.7],
journal  [-7.0], \\
\textit{A \underline{\textbf{book}} has pages, is made of paper, has a cover, is read, has words, is found in libraries, is used for pleasure, has pictures, has information, and has a spine.} &  \textbf{book  [-0.0]},
novel  [-4.9],
manuscript  [-5.4],
journal  [-5.4],
dictionary  [-5.9] &
\textbf{book  [-0.0]},
novel  [-4.3],
dictionary  [-6.1],
paperback  [-6.3],
journal  [-6.4]\\
\midrule
\textit{A \underline{\textbf{helicopter}} flies.} &                           moth [-1.8], bird [-2.3], fly [-2.7], crow [-3.0], bee [-3.0] &                       bird [-2.2], bee [-2.4], butterfly [-2.7], bat [-2.9], moth [-3.0] \\
\textit{A \underline{\textbf{helicopter}} flies, is made of metal, has rotors, has a pilot, and is noisy.} &       \textbf{helicopter} [-.9], bird [-3.1], drone [-3.3], plane [-3.8], rotor [-3.9] &       plane [-.2], \textbf{helicopter} [-2.1], bird [-4.5], jet [-4.8], airplane [-5.9] \\
\textit{A \underline{\textbf{helicopter}} flies, is made of metal, has rotors, has a pilot, is noisy, has blades, has a propeller, is a form of transport, has an engine, and carries people.} &       \textbf{helicopter} [-.3], plane [-3.0], bird [-3.7], vehicle [-4.2], car [-4.5] &  plane [-.2], \textbf{helicopter} [-2.1], bird [-4.7], airplane [-5.5], aircraft [-5.8] \\
\midrule
\textit{A \underline{\textbf{taxi}} is expensive.} &                      car [-2.5], house [-3.5], divorce [-4.1], ticket [-4.1], horse [-4.7] &                              car [-3.0], house [-4.0], lot [-4.1], life [-4.5], horse [-4.6] \\
\textit{A \underline{\textbf{taxi}} is expensive, is yellow, is black, is a car, and is for transport.} &                       car [-1.0], bicycle [-3.0], vehicle [-3.4], horse [-3.5], bus [-4.1] &             Mercedes [-1.8], \textbf{taxi} [-1.9], bus [-2.4], Bentley [-3.0], Jaguar [-3.0] \\
\textit{A \underline{\textbf{taxi}} is expensive, is yellow, is black, is a car, is for transport, is made of metal, has a meter, has wheels, has passengers, and is useful.} &             car [-.7], bicycle [-2.3], vehicle [-3.0], horse [-3.8], \textbf{taxi} [-4.2] &                   \textbf{taxi} [-1.3], bus [-1.5], car [-2.0], bicycle [-2.8], train [-3.5] \\
\midrule
\textit{A \underline{\textbf{telephone}} is made of plastic.} &                        shield [-4.4], chair [-4.4], helmet [-4.5], mask [-4.7], cap [-4.7] &                          car [-3.1], condom [-3.7], banana [-3.7], toy [-3.8], toilet [-4.0] \\
\textit{A \underline{\textbf{telephone}} is made of plastic, is used for communication, has a speaker, rings, and allows you to make calls.} &       phone [-.5], \textbf{telephone} [-1.2], mobile [-4.6], receiver [-4.8], cell [-5.1] &            phone [-.4], \textbf{telephone} [-1.8], bell [-5.0], radio [-5.3], mobile [-5.3] \\
\textit{A \underline{\textbf{telephone}} is made of plastic, is used for communication, has a speaker, rings, allows you to make calls, has a receiver, has a wire, is mobile, has buttons, and has a dial.} &      phone [-.8], \textbf{telephone} [-.8], radio [-4.6], mobile [-4.9], receiver [-5.3] &            phone [-.6], \textbf{telephone} [-1.3], radio [-5.2], mobile [-6.1], bell [-6.8] \\
    \bottomrule
    \end{tabular}
    \caption{Examples of models' predicted completions with 1, 5, and 10 `clue' features provided. Associated log probability included in square brackets.}
    \label{tab:feature2concept-qualitative}
\end{figure*}

%% file: tables/concept2feature_failures.tex
\begin{figure*}[t!]
    \centering
    \small
    \begin{tabular}{p{.2\textwidth}|p{.125\textwidth}|p{.125\textwidth}|p{.125\textwidth}|p{.125\textwidth}|p{.125\textwidth}}
    \toprule
    \textbf{Context} & \textbf{Human}& \proc{BERT-B} & \proc{BERT-L} & \proc{RoBERTa-B} & \proc{RoBERTa-L}  \\ \midrule
    \textit{Everyone knows that a \textbf{hamster} is \blank.}& 
    \textit{small, alive, cute, white, black}& 
    \textit{dangerous, good, right, funny} & 
    \textit{dead, real, dangerous, involved} & 
    \textit{dangerous, evil, bad, dead} & 
    \textit{cute, adorable, harmless, alive} \\
    \midrule
    \textit{Everyone knows that a \textbf{bucket} is a \blank.}& 
    \textit{container, vessel, cylinder}& 
    \textit{bucket, toilet, problem, mess} & 
    \textit{bucket, toilet, weapon, tank} & 
    \textit{bucket, toilet, bomb, hat} & 
    \textit{toilet, bucket, tool, container} \\
    \midrule
    \textit{Everyone knows that a \textbf{motorcycle} has \blank.}& 
    \textit{wheels, seats, lights, brakes, gears}& 
    \textit{wheels, arrived, escaped, tires} & 
    \textit{crashed, arrived, died, power} & 
    \textit{legs, wings, wheels, power} & 
    \textit{wheels, brakes, horsepower, power} \\
    \midrule
    \textit{Everyone knows that an \textbf{anchor} has a \blank.}& 
    \textit{chain, cable, rope, point}& 
    \textit{problem, story, weakness, camera} & 
    \textit{purpose, life, weakness, soul} & 
    \textit{point, voice, pulse, personality} & 
    \textit{job, personality, voice, story} \\
    \midrule
    \textit{Everyone knows that a \textbf{sock} is made of \blank.}& 
    \textit{cotton, fabric, cloth, material, wool}& 
    \textit{wood, leather, steel, iron, metal} & 
    \textit{cotton, rubber, wool, leather, plastic} & 
    \textit{rubber, wood, metal, plastic, bones} & 
    \textit{cotton, wool, fabric, rubber, material} \\
    \bottomrule
    \end{tabular}
    \caption{Examples of models' predicted completions to concept/relation prompts targeting the production of properties. Predictions are over the full vocabulary intersection.}
    \label{tab:concept2feature-failures}

\end{figure*}

%% file: tables/pasca_attributes.tex
\begin{figure*}[!t]
    \centering
    \small
    \begin{tabular}{p{.35\textwidth}|p{.25\textwidth}p{.25\textwidth}} 
    \toprule
    \multirow{2}*{\textbf{\citeA{pasca_you_2007} Example} }   & \multicolumn{2}{c}{\proc{Roberta-L}}\\
    & \textbf{Concept from Attributes} &  \textbf{Attributes from Concept}\\ \midrule \midrule
A \textbf{\textul{company}} has a \textbf{CEO}, a \textbf{future},
a \textbf{president}, a \textbf{competitors},
a \textbf{mission statement}, an \textbf{owner},
a \textbf{website}, an \textbf{organizational structure},
a \textbf{logo}, and a \textbf{market share}.
& \textbf{company [0.695]} , business [0.23], corporation [0.03], startup [0.02], brand [0.01]  
& CEO [0.15], culture [0.1], mission [0.04], price [0.03], hierarchy [0.03],  strategy [0.03]
\\ \midrule
A \textbf{\textul{country}} has a \textbf{capital}, a \textbf{population}
a \textbf{president}, a \textbf{map},
a \textbf{capital city}, a \textbf{currency},
a \textbf{climate}, a \textbf{flag},
a \textbf{culture}, and a \textbf{leader}.
&  \textbf{country [0.72]}, nation [0.25], state [0.03], republic [0.002], government [0.001] & 
constitution [0.23], history [0.07], culture [0.07], soul [0.04], budget [0.03], border [0.03]\\ \midrule
A \textbf{\textul{drug}} has
a \textbf{side effect}, a \textbf{cost},
\textbf{structure}, a \textbf{benefit},
a \textbf{mechanism}, \textbf{overdose},
\textbf{use}, a \textbf{price},
and a \textbf{pharmacology}. 
& \textbf{drug [0.9]}, medicine [0.02], product [0.02], medication [0.02], substance [0.01]
&  effect [0.1], risk [0.1], dependency [0.06], potential [0.05], cost [0.04]\\ \midrule
A \textbf{\textul{painter}} has
\textbf{paintings}, \textbf{works},
a \textbf{portrait}, a \textbf{death},
a \textbf{style}, a \textbf{artwork},
a \textbf{bibliography}, a \textbf{bio},
and a \textbf{childhood}.
& person [0.21], \textbf{painter [0.2]}, writer [0.14], poet [0.05], book [0.04]
&  style [0.15], voice [0.1], vision [0.07], technique [0.03], palette [0.03], soul [0.03]\\
    \bottomrule
    \end{tabular}
    \caption{\proc{RoBERTa-L} captures the concept/attribute pairs automatically extracted by Pasca \& Van Durme (2007) based on web log frequency (target completions bolded), as shown by predictions of both concept and properties (associated probability in brackets).}
    \label{fig:pasca}
\end{figure*}